\documentclass{IOS-Book-Article}

\usepackage{times}
\usepackage{xcolor}
\usepackage{soul}
\usepackage[utf8]{inputenc}
\usepackage[small]{caption}

\usepackage{amsthm}
\usepackage{amsmath}
\usepackage{amssymb}
\usepackage{graphicx}

\usepackage{color, colortbl}
\definecolor{LightGray}{gray}{0.9}

\newtheorem{theorem}{Theorem}
\newtheorem{proposition}{Proposition}
\newtheorem{corollary}{Corollary}
\newtheorem{definition}{Definition}
\newtheorem{transformation}{Transformation}

\def\hb{\hbox to 10.7 cm{}}

\begin{document}

\pagestyle{headings}
\def\thepage{}

\begin{frontmatter}              

\title{A Preliminary Report on Probabilistic Attack Normal Form for Constellation Semantics}


\author[A,B]{Theofrastos Mantadelis\thanks{Project: TBD}}
and
\author[A,B]{Stefano Bistarelli}

\address[A]{Department of Mathematics and Computer Science, University of Perugia, Italy}
\address[B]{Member of the INdAM Research group GNCS}

\begin{abstract}
After Dung's founding work in Abstract Argumentation Frameworks there has been a growing interest in extending the Dung's semantics in order to describe more complex or real life situations. Several of these approaches take the direction of weighted or probabilistic extensions. One of the most prominent probabilistic approaches is that of constellation Probabilistic Abstract Argumentation Frameworks from Li~et~al.

In this paper, we present a normal form for constellation probabilistic abstract argumentation frameworks. Furthermore, we present a transformation from general constellation probabilistic abstract argumentation frameworks to the presented normal form. In this way we illustrate that the simpler normal form has equal representation power with the general one.
\end{abstract}

\end{frontmatter}

\section{Introduction}
Argumentation is an everyday method of humanity to discuss and solve myriad different situations where opinions or point of views conflict. Abstract Argumentation Frameworks~\cite{DBLP:journals/jlp/Dung95} (AAFs) aim in modeling everyday situations where information is inconsistent or incomplete. Many different extensions of AAFs from Dung's pioneering work have appeared in order to describe different everyday situations. Sample works includes assumption based argumentation~\cite{Bondarenko:1993:AFN:164431.164472}, extending AAFs with support~\cite{DBLP:conf/comma/OrenN08}, introducing labels~\cite{Wu2010ALJ}. Other approaches of extending AAFs have focused on introducing weights in elements of the AAF, such as~\cite{DBLP:journals/logcom/Bench-Capon03,DBLP:journals/ijar/BistarelliRS18}. Such approaches are powerful tools to model voting systems, belief in arguments and argument strength.

Knowledge representation with the use of probabilistic information has been used in many areas of computer science. Probabilistic information, is a powerful medium to represent knowledge. Similarly, many researchers have extended AAFs by adding probabilistic information. These very prominent extensions of AAFs have been categorized in two big groups by Hunter~\cite{DBLP:conf/comma/Hunter12}: the \textbf{epistemic} approaches and the \textbf{constellation} approaches.

The epistemic approaches, such as those presented in~\cite{DBLP:conf/ecai/Thimm12,DBLP:journals/jair/HunterT17} describe probabilistic AAFs that the uncertainty does not alter the structure of the AAFs. These type of AAFs use the probability assignments to quantify the existing uncertainty of arguments in AAFs and not to introduce new uncertainty.

The constellation approaches, such as those presented in~\cite{DBLP:conf/tafa/LiON11,DBLP:conf/ijcai/FazzingaFP13,DBLP:conf/sum/DoderW14,DBLP:journals/cas/Dondio14,DBLP:journals/corr/LiaoXH16} introduce probabilistic elements in the AAF in such a way that the structure of the AAF becomes uncertain. The constellation approaches generate a set of AAFs with a probabilistic distribution and as such define a probabilistic distribution over the extensions of those AAFs.

In this paper we focus on the constellation approach from Li~et~al.~\cite{DBLP:conf/tafa/LiON11}. \cite{DBLP:conf/tafa/LiON11} introduced probabilistic elements to the structure of AAFs, resulting to a set of AAFs. This allows for a set of arguments to be an (admissible, stable, ground, etc.) extension in some of the AAFs that are represented by the constellation. This simple but yet powerful representation has the ability to represent naturally many different uncertain scenarios.

The works of~\cite{DBLP:conf/tafa/LiON11}~and~\cite{DBLP:conf/comma/Hunter12} can be considered as the pioneering work on combining probabilities with AAFs. Our motivation for this work is to present a normal form for PrAAFs that while simpler has the same expression power with general PrAAFs; we also present a PrAAF to PrAAF transformation that allows any general PrAAF to be converted to an equivalent PrAAF in normal form.


\section{Preliminaries}
\label{sec:Preliminaries}
\subsection{Abstract Argumentation}
\begin{figure}[htbp]
\vspace{-0.2cm}
\centering
\includegraphics[width=0.35\textwidth]{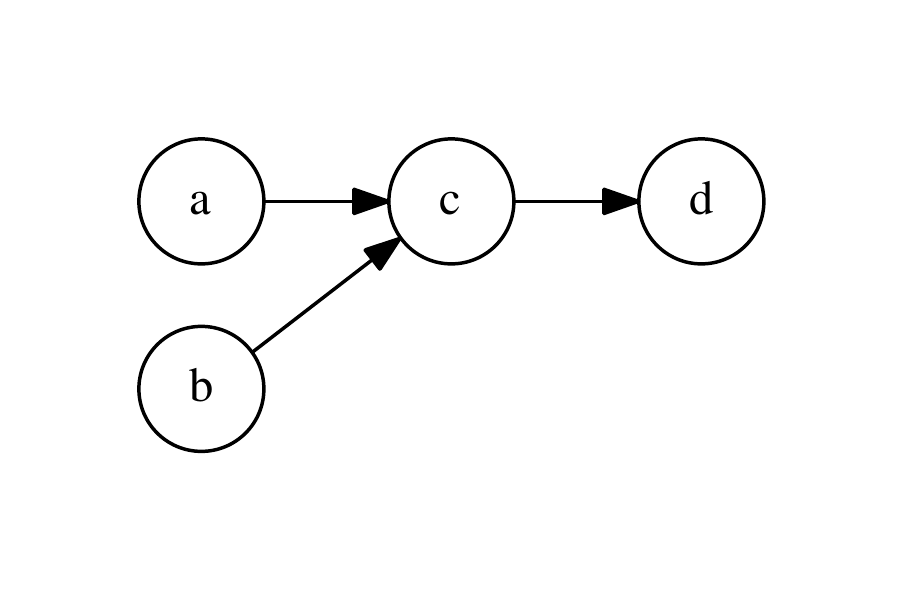}
\vspace{-0.2cm}
\caption{Exmaple AAF $(\{a,b,c,d\}, \{a\rightarrow c, b \rightarrow c, c \rightarrow d\})$. Arguments are represented as cycles and attacks as arrows. Arguments $a$, $b$ are attacking argument $c$ which attacks argument $d$.} 
\label{fig:AAF}
\end{figure}

An abstract argumentation framework~\cite{DBLP:journals/jlp/Dung95} (AAF) is a tuple $AAF = (Args, Atts)$ where $Args$ is a set of arguments and $Atts$ a set of attacks among arguments of the form of a binary relation $Atts \subseteq Args \times Args$. For arguments $a, b \in Args$, we use $a \rightarrow b$ as a shorthand to indicate $(a, b)\in Atts$ and we say that argument $a$ \textit{attacks} argument $b$. Figure~\ref{fig:AAF} illustrates an example AAF.

A set of arguments $S \subseteq Args$ is said to be \emph{conflict-free} iff $\nexists a, b \in S$ where $a \rightarrow b \in Atts$. An argument $a \in Args$ is acceptable with respect to set $S \subseteq Args$ if no argument attack $a$ or if  $\forall b \in Args$ that $\exists b \rightarrow a \in Atts$ then $\exists c \in S$ where $c \rightarrow b \in Atts$.

Given the above~\cite{DBLP:journals/jlp/Dung95} gives semantics to AAF by the use of extensions over subsets of arguments. Dung first defines the \emph{admissible} semantics. A set $S \subseteq Args$ is admissible iff $S$ is conflict free and each $a \in S$ is acceptable with respect to $S$. Following our example AAF from Figure~\ref{fig:AAF}, the set $\{a,b,d\}$ is admissible. Over time several different semantics have been discussed such as complete, preferred, grounded, stable~\cite{DBLP:journals/jlp/Dung95}, semi-stable~\cite{doi:10.1093/logcom/exr033}, CF2~\cite{Gaggl:2010:CSR:1860828.1860852} etc.

\subsection{Constellation based Probabilistic Abstract Argumentation Frameworks}
Hunter~\cite{DBLP:conf/comma/Hunter12}, categorizes probabilistic abstract argumentation frameworks (PrAAFs) in two different categories: the \emph{constellation} and the \emph{epistemic} PrAAFs. For this paper we will focus on the constellation approaches and we base our work in the definition of PrAAFs by~\cite{DBLP:conf/tafa/LiON11}.

A constellation approach to PrAAFs defines probabilities over the structure of the AAF graph. One can assign probabilities to either the arguments or/and attacks of the AAF. We refer to arguments/attacks with assigned probabilities less than $1$ as probabilistic arguments/attacks and we refer as probabilistic elements to either probabilistic arguments or probabilistic attacks.

A probabilistic elment $e$ exists in an AAF with probability $P(e)$. These probabilistic elements correspond to random variables, which are assumed to be mutually independent\footnote{As we are going to present later in the paper, the structure of AAF might impose dependencies among otherwise assumed independent probabilistic elements.}. As such, a PrAAF defines a probability distribution over a set of AAFs.

\begin{figure}[htbp]
\vspace{-0.2cm}
\centering
\includegraphics[width=0.35\textwidth]{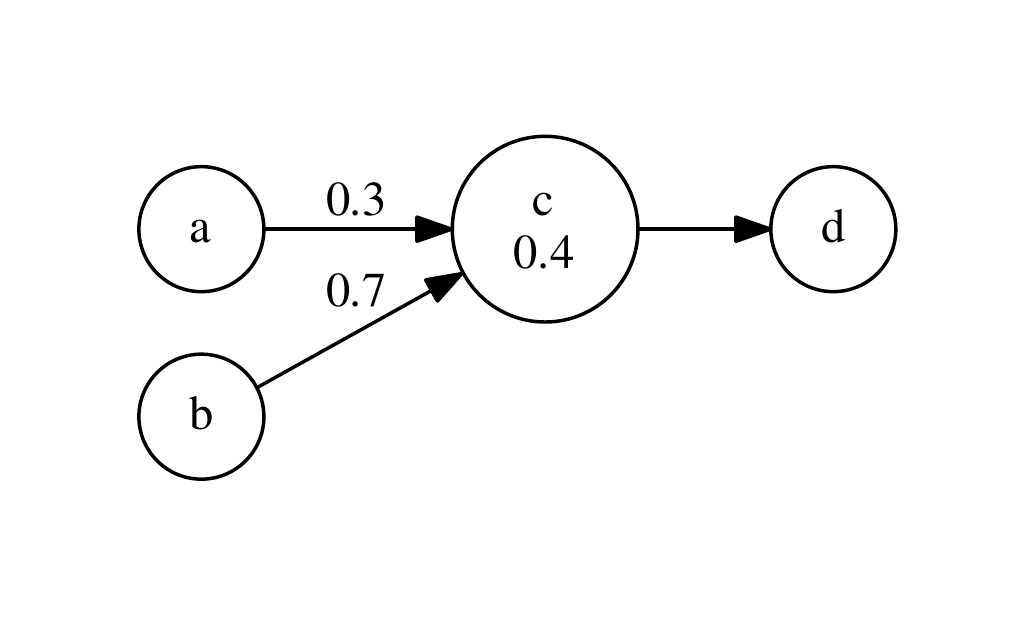}
\vspace{-0.2cm}
\caption{Example PrAAF $(\{a,b,c,d\}, \{1,1,0.4,1\}, \{a\rightarrow c, b \rightarrow c, c \rightarrow d\},\{0.3,0.7,1\})$. Arguments are represented as cycles and attacks as arrows.}
\label{fig:PrAAF}
\end{figure}
\begin{definition}\label{def:PrAAF}
Formally, a PrAAF is a tuple $PrAAF = (Args, P_{Args}, Atts, P_{Atts})$ where $Args$, $Atts$ define an AAF, $P_{Args}$ is a set of probabilities for each $a \in Args$ with $0 < P_{Args}(a) \leq 1$ and $P_{Atts}$ is a set of probabilities for each $\rightarrow \in Atts$ with $0 < P_{Atts}(\rightarrow) \leq 1$.
\end{definition}
Finally, stating an argument or an attack having probability $0$ is redundant. A probabilistic argument or attack with $0$ probability is an argument or attack that is not part of any AAF that the constellation represents. Figure~\ref{fig:PrAAF}, illustrates an example PrAAF with 3 different probabilistic elements.

\begin{table*}[ht]
\centering
\begin{tabular}{cp{0.21\textwidth}ll}
\textbf{AAF} & \textbf{Possible World} & \textbf{Prob.} & \textbf{Admissible Sets} \\\hline
\rowcolor{LightGray}
\raisebox{-0.5\totalheight}{\includegraphics[trim=35 30 35 30,clip,width=0.16\textwidth]{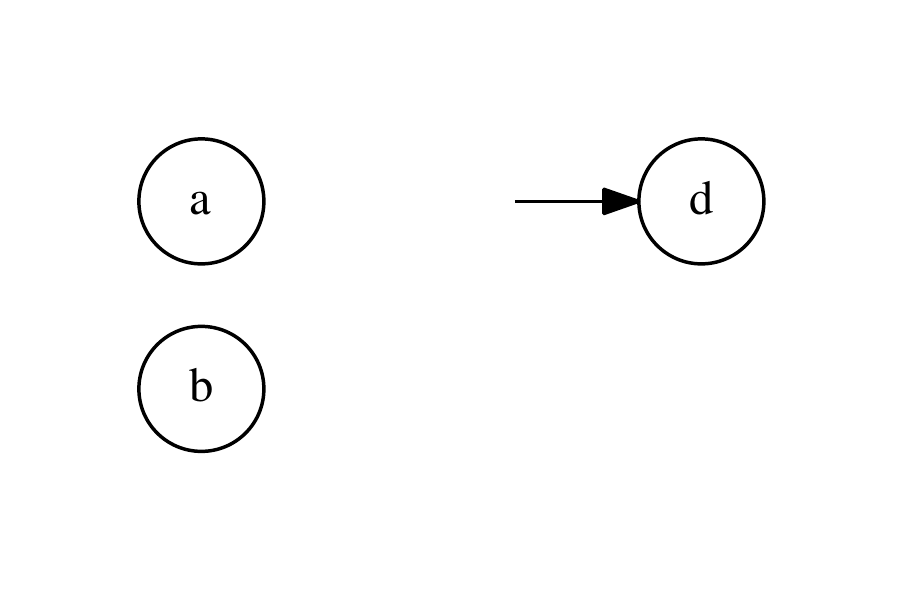}} & $\lnot (a \rightarrow c) \land \lnot (b \rightarrow c) \land \lnot c $ & 0.126 & $\{\}, \{a\}, \{b\}, \{d\}, \{a,b\}, \{a,d\}, \{b,d\}, \{a,b,d\}$ \\
\raisebox{-0.5\totalheight}{\includegraphics[trim=35 30 35 30,clip,width=0.16\textwidth]{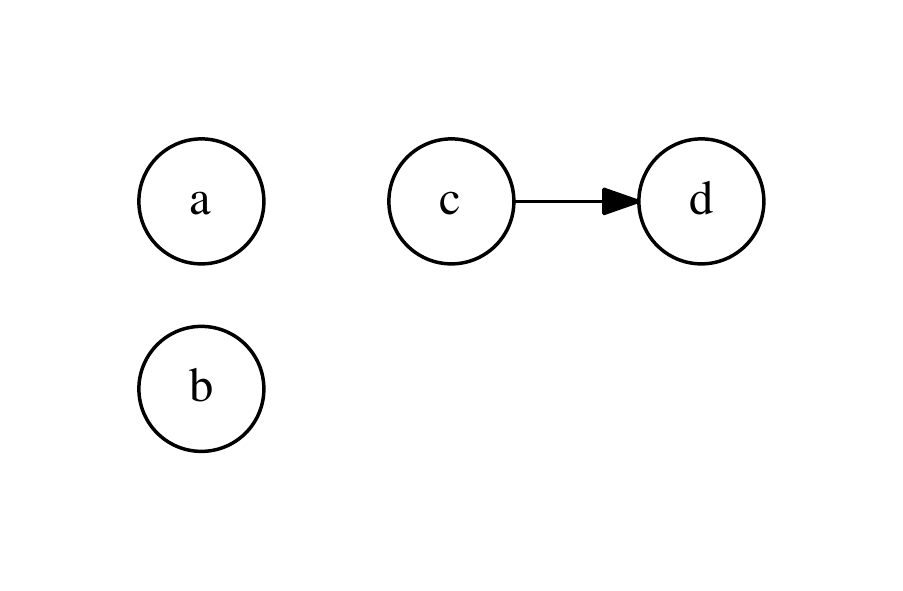}} & $\lnot (a \rightarrow c) \land \lnot (b \rightarrow c) \land c $       & 0.084 & $\{\}, \{a\}, \{b\}, \{c\}, \{a,b\}, \{a,c\}, \{b,c\}, \{a,b,c\}$ \\
\rowcolor{LightGray}
\raisebox{-0.5\totalheight}{\includegraphics[trim=35 30 35 30,clip,width=0.16\textwidth]{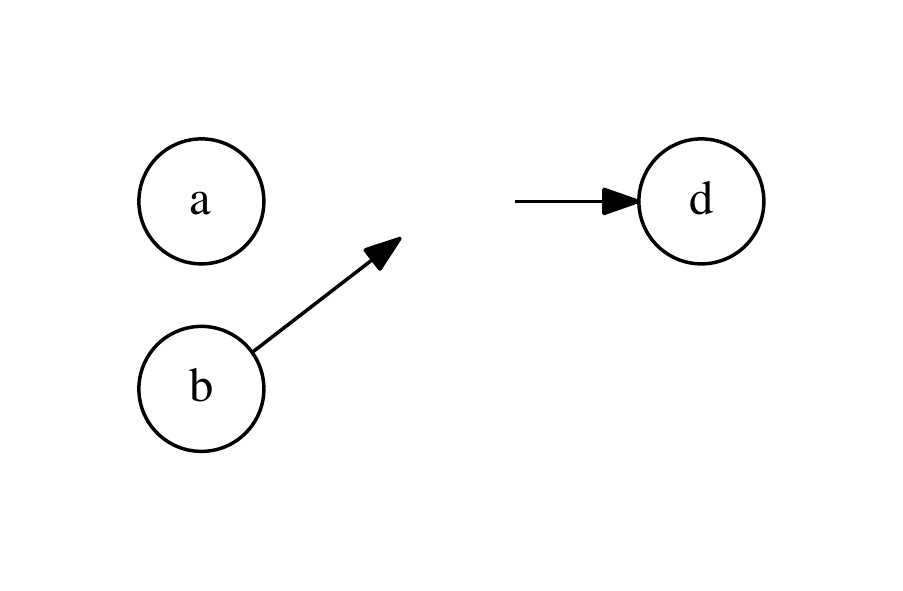}} & $\lnot (a \rightarrow c) \land (b \rightarrow c) \land \lnot c $       & 0.294  & $\{\}, \{a\}, \{b\}, \{d\}, \{a,b\}, \{a,d\}, \{b,d\}, \{a,b,d\}$\\
\raisebox{-0.5\totalheight}{\includegraphics[trim=35 30 35 30,clip,width=0.16\textwidth]{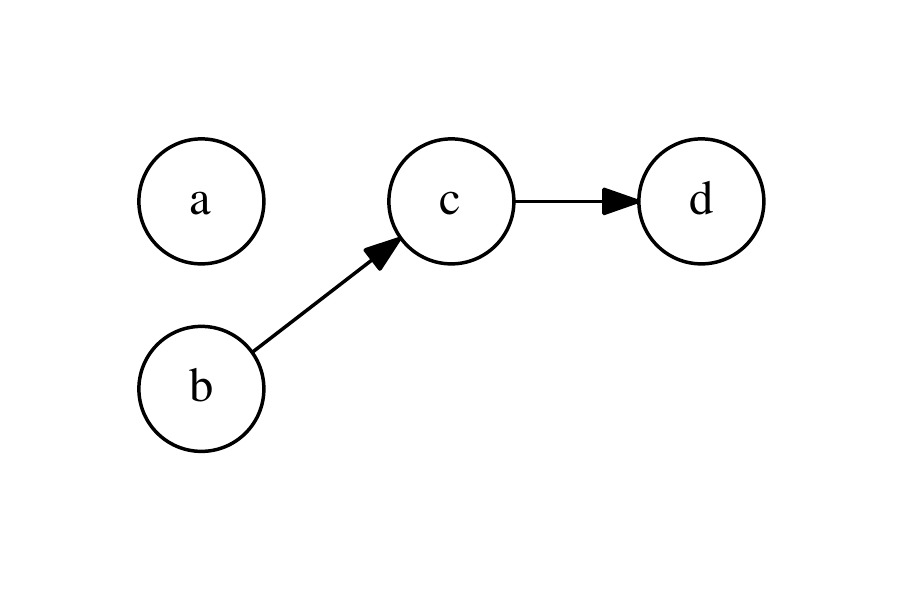}} & $\lnot (a \rightarrow c) \land (b \rightarrow c) \land c $             & 0.196  & $\{\}, \{a\}, \{b\}, \{a,b\}, \{b,d\}, \{a,b,d\}$\\
\rowcolor{LightGray}
\raisebox{-0.5\totalheight}{\includegraphics[trim=35 30 35 30,clip,width=0.16\textwidth]{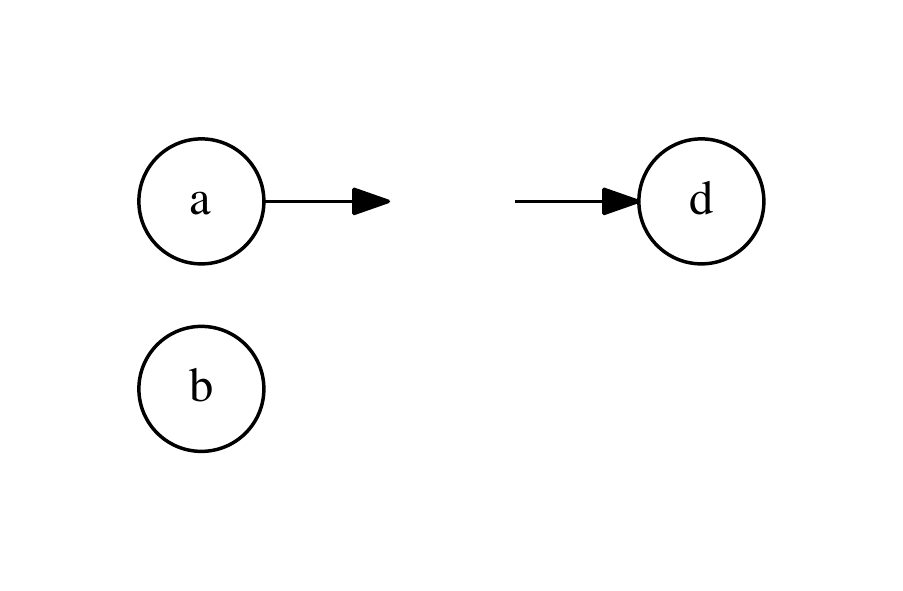}} & $(a \rightarrow c) \land \lnot (b \rightarrow c) \land \lnot c $       & 0.054  & $\{\}, \{a\}, \{b\}, \{d\}, \{a,b\}, \{a,d\}, \{b,d\}, \{a,b,d\}$\\
\raisebox{-0.5\totalheight}{\includegraphics[trim=35 30 35 30,clip,width=0.16\textwidth]{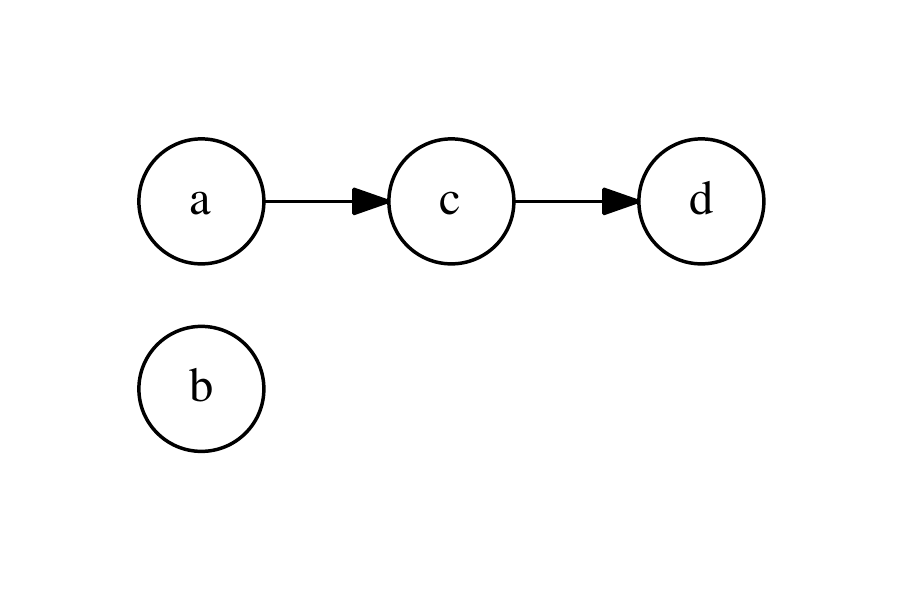}} & $(a \rightarrow c) \land \lnot (b \rightarrow c) \land c $                   & 0.036  & $\{\}, \{a\}, \{b\}, \{a,b\}, \{a,d\}, \{a,b,d\}$\\
\rowcolor{LightGray}
\raisebox{-0.5\totalheight}{\includegraphics[trim=35 30 35 30,clip,width=0.16\textwidth]{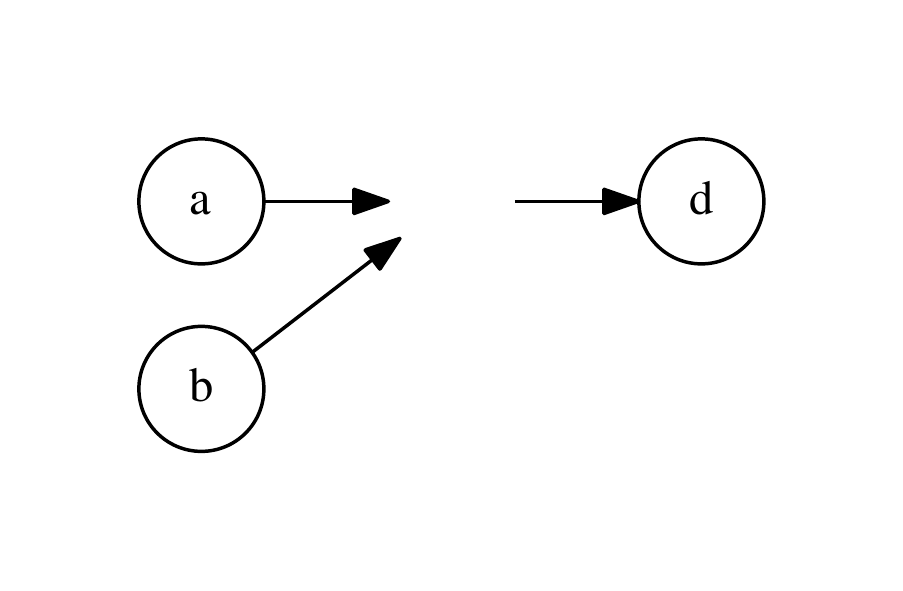}} & $(a \rightarrow c) \land (b \rightarrow c) \land \lnot c $             & 0.126  & $\{\}, \{a\}, \{b\}, \{d\}, \{a,b\}, \{a,d\}, \{b,d\}, \{a,b,d\}$\\
\raisebox{-0.5\totalheight}{\includegraphics[trim=35 30 35 30,clip,width=0.16\textwidth]{AAF.pdf}}         & $(a \rightarrow c) \land (b \rightarrow c) \land c $                   & 0.084  & $\{\}, \{a\}, \{b\}, \{a,b\}, \{a,d\}, \{b,d\}, \{a,b,d\}$\\\hline
\end{tabular}
\caption{Possible worlds of our example PrAAF from Figure~\ref{fig:PrAAF}. Shaded rows, illustrate possible worlds that generate an invalid AAF.}
\label{tbl:possible_worlds}
\end{table*}

As mentioned a PrAAF defines a probability distribution for all the possible non-probabilistic AAFs it contains. Each single possible set of probabilistic elements (arguments or attacks) of the PrAAF is called a \textbf{possible world}. Table~\ref{tbl:possible_worlds} presents all possible worlds for the example PrAAF of Figure~\ref{fig:PrAAF}. One can notice that having only three different probabilistic elements it generates eight possible worlds. The possible worlds of a PrAAF are exponential in the number of probabilistic elements ($2^N$ where $N$ the number of probabilistic elements).
\begin{definition}[Probability of Possible World]
The probability of a possible world equals to the product of the probability of each probabilistic element that is in the possible world with the product of one minus the probability of each probabilistic element that is excluded from the possible world.
\begin{equation*}
P_{world} = \prod\limits_{e_i\in AAF_{world}}P(e_i) \cdot \prod\limits_{e_j \notin AAF_{world}}(1-P(e_j))
\end{equation*}
\end{definition}

There has been extensive research on how to compute the probabilities without explicitly enumerating all possible worlds. We point out some works from Probabilistic Logic Programming that is closely related with PrAAFs. For efficient exact inference we direct the reader at~\cite{MANTADELIS10ICLP,RIGUZZI11CORR,KIMMIG11TPLP,FIERENS2013TPLP,SHTERIONOV15PADL}, and for efficient approximate inference at~\cite{BRAGAGLIA11ILP,VLASSELAER15IJCAI,DBLP:conf/padl/MantadelisR17}.

\subsection{Imposing Restrictions}
Li~et~al.~\cite{DBLP:conf/tafa/LiON11} further restrict the possible worlds to what is called induced AAFs. Furthermore, states that the probabilities $P_{atts}$ are conditional probabilites instead the likelihood of existence. These restrictions appear as a separate definition, formally:
\begin{definition}[Inducing an AAF from a PrAAF] An AAF $(Args_{Ind}, Atts_{Ind})$ is said to be induced from a PrAAF $(Args, P_{Args}, Atts, P_{Atts})$ iff all of the following hold:
\begin{enumerate}
  \item $Args_{Ind} \subseteq Args$
  \item $Atts_{Ind} \subseteq Atts \cap (Args_{Ind} \times Args_{Ind})$
  \item $\forall a\in Args$ such that $ P_{Args}(a) = 1, a \in Args_{Ind}$
  \item $\forall a1\rightarrow a2\in Atts$ such that $P_{Atts}(a1\rightarrow a2) = 1$ and $ P_{Args}(a1) = P_{Args}(a2) = 1, a1\rightarrow a2 \in Atts_{Ind}$
\end{enumerate}
\end{definition}
Furthermore, $P_{Att}(a1\rightarrow a2)$ is stated to be the conditional probability of the attack existing when both attacking and attacked argument exist in the AAF ($P_{Att}(a1\rightarrow a2|a1,a2\in Args_{Ind})$).

\section{Normalizing Constellation Approaches}
\label{sec:Normalize}

\subsection{PrAAF Normal Forms}
A simpler approach to restric the possible worlds without introducing extra restrictions is to restrict PrAAFs only to probabilistic attacks and not probabilistic arguments.
\begin{definition}[Probabilistic Attack Normal Form]
A PrAAF $P$ is in its probabilistic attack normal form if it contains no probabilistic arguments ($\forall a \in Args, P(a) = 1$).
\end{definition}

As we illustrate later in the paper any PrAAF can be transformed to a normal form PrAAF that has equivalent possible worlds and define equivalent probability distribution over their admissible extensions. Thus we illustrate that normal form PrAAFs are both the simplest but yet sufficient forms for representing any PrAAF.

\section{Transforming General PrAAFs to Normal Form PrAAFs}
In order to motivate our normal form for PrAAFs, we illustrate that any general PrAAF can be represented to an equivalent normal form PrAAF. To do so, we present a transformation from any general PrAAF to an equivalent normal form PrAAF.

Our transformation proves, that we can represent as a normal form PrAAF an equivalent constellation of AAFs with the exact same probabilistic distribution over their extensions that a general PrAAF would represent. This indicates that even with our simpler definition for PrAAFs we can model all the same PrAAFs.

\subsection{Transforming Probabilistic Arguments to Probabilistic Attacks}
We present a transformation of probabilistic argument to probabilistic attack. We start by defining a special argument called \emph{ground truth}:
\begin{definition}[Ground Truth]
We introduce a special argument called \emph{Ground Truth} and shorthand it with the letter $\eta$. We say that $\eta$ is undeniably true meaning that $\eta$ is never attacked by any argument and is always included in all extensions regardless the semantics used.
\end{definition}
The $\eta$ argument restricts the extensions allowed by all semantics in such a way that $\eta$ is always included. For example, in the admissible semantics of an AAF without $\eta$ a valid extension is the empty set($\{\}$), but in an AAF that includes $\eta$ the empty set is not a valid extension under the admissible semantics and the equivalent extension is $\{\eta\}$.
\begin{definition}[Acceptable Extensions]
For AAFs that contain $\eta$, an acceptable extension $E$ is one that includes $\eta$ ($\eta \in E$).
\end{definition}
\begin{transformation}[Probabilistic argument to probabilistic attack]
Any PrAAF $P$, can be transformed\footnote{Such transformation is categorized as a normal expansion~\cite{DBLP:conf/comma/BaumannB10} of the original PrAAF.} to an equivalent PrAAF $P'$ by removing any probabilistic information attached to an argument $a \in Args$, with $P(a)$ and introducing a probabilistic attack from the ground truth $\eta$ to argument $a$ with probability $1-P(a)$.
\end{transformation}

\begin{definition}
We notate $P \equiv^{\sigma}_{|\eta} P'$ the standard equivalence~\cite{OIKARINEN2011} of PrAAF $P$ with PrAAF $P'$ under semantics $\sigma$ by ignoring the existence of $\eta$ in the acceptable extensions.
\end{definition}
\begin{theorem}[Equivalence of transformed PrAAF]
A transformed PrAAF $P'$ has an equivalent distribution in terms of admissible sets containing $\eta$ compared with the admissible sets of the original PrAAF $P$.
\end{theorem}
\begin{proof}
We split the proof in two parts. First we show that PrAAF $P$ generates AAFs that have the same admissible sets with the generated AAFs from PrAAF $P'$. We point out that for PrAAF $P'$ acceptable admissible sets are only the ones that contain the ground truth argument which we ignore its existence when comparing admissible sets. For example, the empty admissible set of $P$ is equivalent with the $\{\eta\}$ admissible set of $P'$. A probabilistic argument $pa$ generates two different sets of AAFs, set $S_1$ where $pa$ exists and $S_2$ where $pa$ does not exist. 

PrAAF $P'$ generates $S'_1$ the equivalent sets of $S_1$ when $\eta \rightarrow pa$ does not exist and the equivalent $S'_2$ sets of $S_2$ when $\eta \rightarrow pa$ exists. When comparing an AAF with $pa$ versus an AAF without $\eta \rightarrow pa$ the only difference is the existence of $\eta$ as we only consider admissible sets that contain it and we ignore its existence in the admissible sets the two graphs are equivalent thus the $S'_1$ sets are equivalent with the $S_1$ sets.

For $S_2$ where $pa$ does not exist, the equivalent $S'_2$ contains AAFs where the argument $pa$ is been attacked by $\eta$ and is not defended by any other argument. Clearly, as $\eta$ is included in every extension that we consider then every attack originating from $pa$ is defended; thus, the AAFs of $S_2$ generate the same admissible extensions with the AAFs of $S'_2$.

Next part is to show that the probability of each extension is the same. The probability that a set is an admissible extension is been computed by the summation of the possible worlds where that set is admissible. As $S_1, S_2$ are equivalent with $S'_1, S'_2$ and produce equivalent AAFs then the possible worlds are equivalent too. The probability of each possible world is also the same as when $pa$ would exist the possible world probability is multiplied by $P(pa)$. In the equivalent case the attack $\eta \rightarrow pa$ does not exist and the possible world probability is multiplied by $1-(1-P(pa))=pa$. Similarly, for the possible worlds that $pa$ does not exist.
\end{proof}
\begin{corollary}
PrAAF $P'$ has equivalent acceptable extensions with PrAAF $P$ for all semantics where acceptability of an argument is necessary for the inclusion of the argument in the extension. Such semantics include: complete, preferred, ground and stable semantics. Similarly, as the probabilistic distributions are equivalent then all acceptable extensions of $P'$ will have equal probability with their equivalent extension from $P$.
\end{corollary}

We also want to note that our transformation is a one-to-one reversible transformation. By one-to-one we mean that the transformation generates a unique equivalent PrAAF from any PrAAF and vice versa, assuming that $\eta$ is known.
\begin{proposition}[Reversibility of the transformation]
The probabilistic argument to probabilistic attack transformation is reversible and creates a one-on-one equivalent PrAAF.
\end{proposition}
\begin{proof}
Obvious, one can reverse the transformation by just following the reverse steps of the transformation.
\end{proof}

\begin{figure}[htbp]
\vspace{-0.2cm}
\centering
\includegraphics[width=0.35\textwidth]{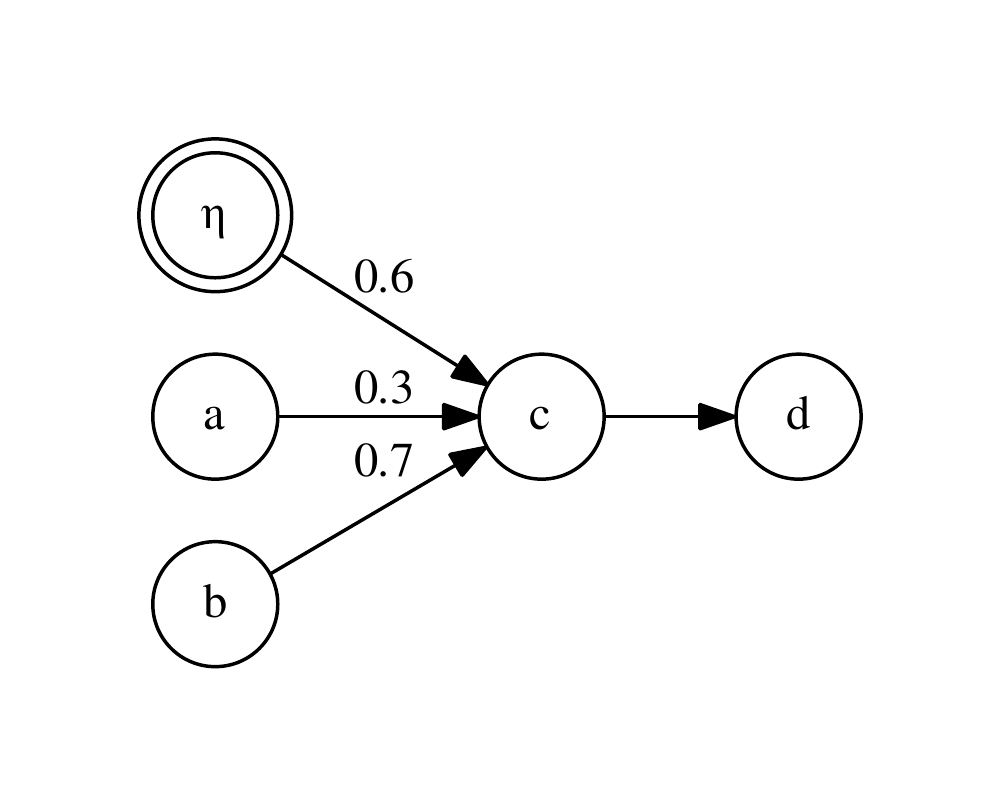}
\vspace{-0.2cm}
\caption{Example transformed PrAAF $(\{a,b,c,d,\eta\},$ $\{1,1,1,1,1\}, \{a\rightarrow c, b \rightarrow c, c \rightarrow d, \eta \rightarrow c\},$ $\{0.3,0.7,1,0.6\})$.}
\label{fig:PrAAF_Can}
\end{figure}
By using the probabilistic argument to probabilistic attack transformation on the PrAAF of Figure~\ref{fig:PrAAF} we get the PrAAF of Figure~\ref{fig:PrAAF_Can}. Table~\ref{tbl:can_possible_worlds} presents the possible worlds of the PrAAF of Figure~\ref{fig:PrAAF_Can}. Now, each possible world represents a valid AAF that generates the equivalent acceptable admissible sets like the original PrAAF. Furthermore, the probabilistic distribution is identical.
\begin{table*}[htbp]
\centering
\begin{tabular}{cp{0.25\textwidth}lp{0.40\textwidth}}
\textbf{AAF} & \textbf{Possible World} & \textbf{Prob.} & \textbf{Acceptable Admissible Sets} \\\hline
\raisebox{-0.5\totalheight}{\includegraphics[trim=35 30 35 30,clip,width=0.16\textwidth]{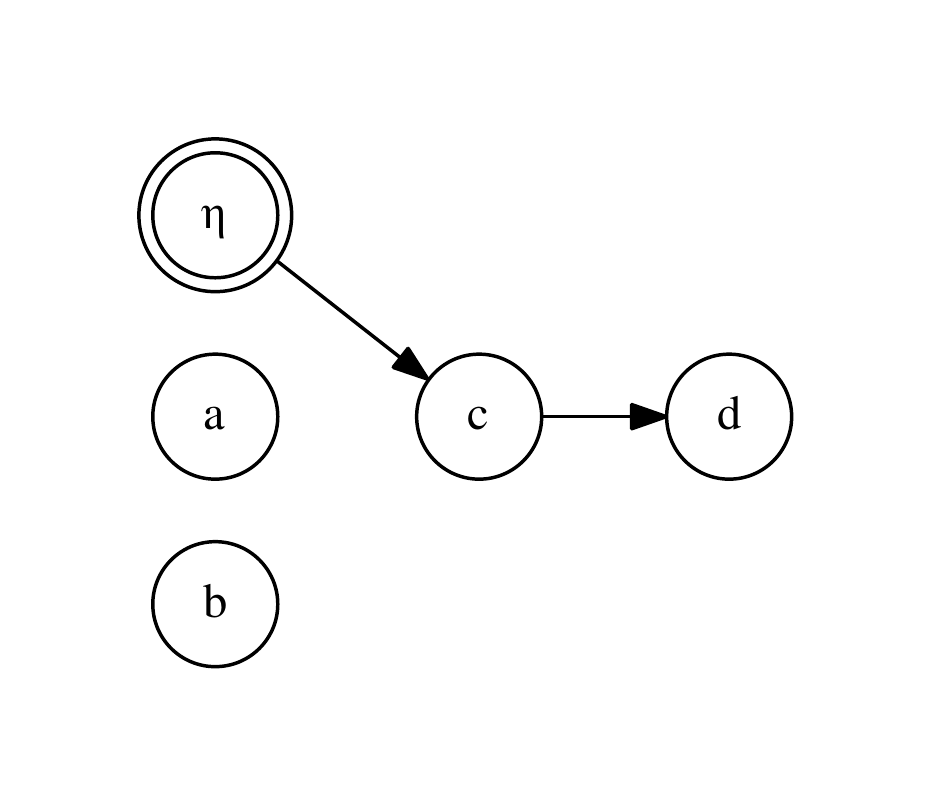}} & $\lnot (a \rightarrow c) \land \lnot (b \rightarrow c) \land (\eta \rightarrow c) \equiv \lnot (a \rightarrow c) \land \lnot (b \rightarrow c) \land \lnot c$       & 0.126  & $\{\eta\}, \{\eta,a\}, \{\eta,b\}, \{\eta,d\}, \{\eta,a,b\},$ $\{\eta,a,d\}, \{\eta,b,d\}, \{\eta,a,b,d\}$\\
\raisebox{-0.5\totalheight}{\includegraphics[trim=35 30 35 30,clip,width=0.16\textwidth]{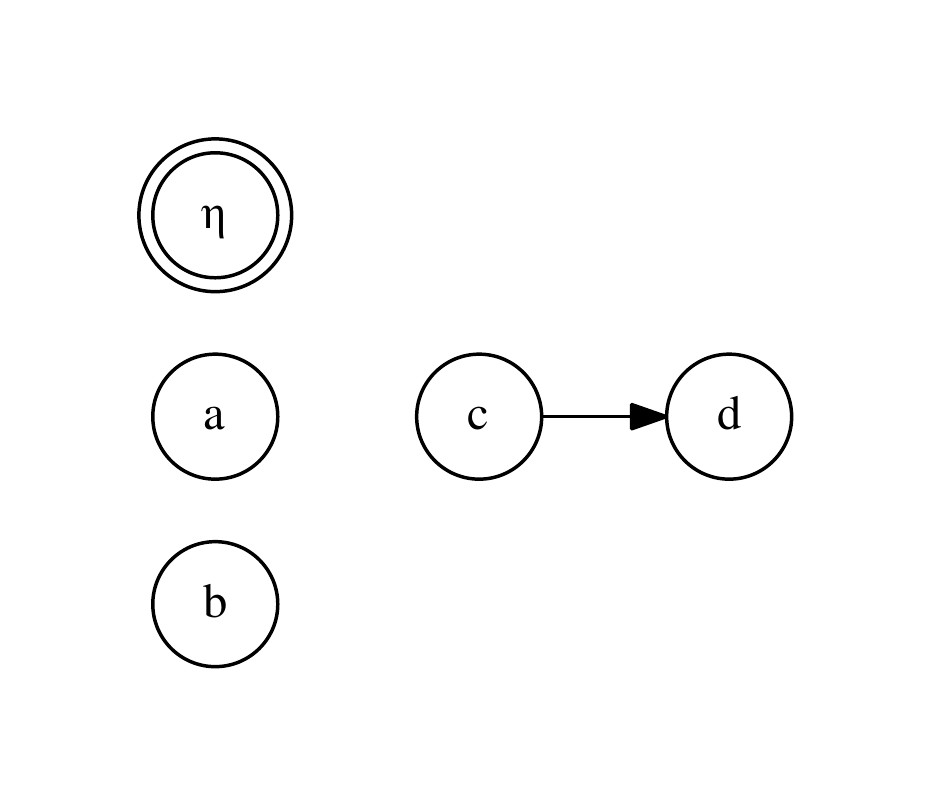}} & $\lnot (a \rightarrow c) \land \lnot (b \rightarrow c) \land \lnot (\eta \rightarrow c) \equiv \lnot (a \rightarrow c) \land \lnot (b \rightarrow c) \land c $ & 0.084   & $\{\eta\}, \{\eta,a\}, \{\eta,b\}, \{\eta,c\}, \{\eta,a,b\},$ $\{\eta,a,c\}, \{\eta,b,c\}, \{\eta,a,b,c\}$\\
\raisebox{-0.5\totalheight}{\includegraphics[trim=35 30 35 30,clip,width=0.16\textwidth]{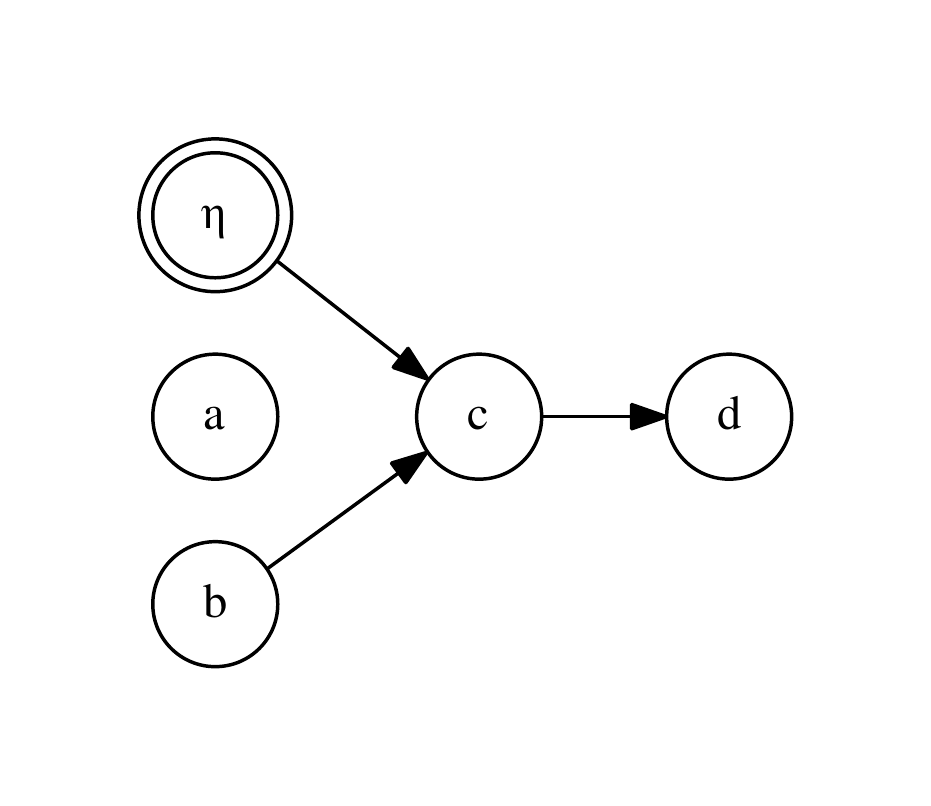}} & $\lnot (a \rightarrow c) \land (b \rightarrow c) \land (\eta \rightarrow c)  \equiv \lnot (a \rightarrow c) \land  (b \rightarrow c) \land \lnot c$             & 0.294 & $\{\eta\}, \{\eta,a\}, \{\eta,b\}, \{\eta,d\}, \{\eta,a,b\},$ $\{\eta,a,d\}, \{\eta,b,d\}, \{\eta,a,b,d\}$\\
\raisebox{-0.5\totalheight}{\includegraphics[trim=35 30 35 30,clip,width=0.16\textwidth]{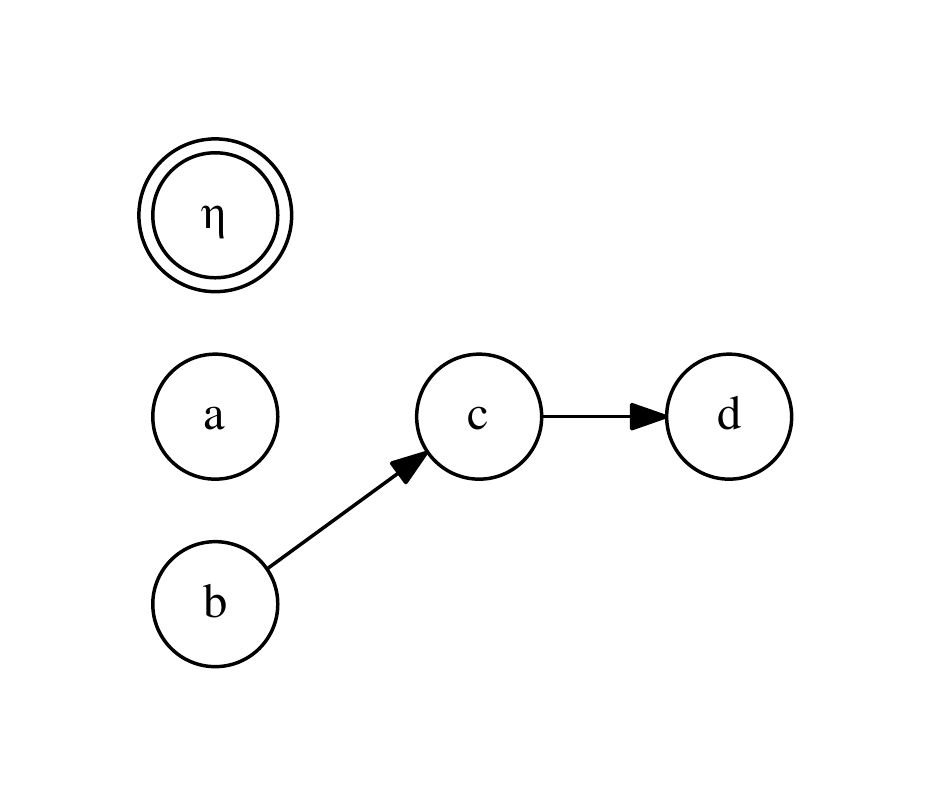}} & $\lnot (a \rightarrow c) \land (b \rightarrow c) \land \lnot (\eta \rightarrow c)  \equiv \lnot (a \rightarrow c) \land  (b \rightarrow c) \land  c$       & 0.196   & $\{\eta\}, \{\eta,a\}, \{\eta,b\}, \{\eta,a,b\}, \{\eta,b,d\},$ $\{\eta,a,b,d\}$\\
\raisebox{-0.5\totalheight}{\includegraphics[trim=35 30 35 30,clip,width=0.16\textwidth]{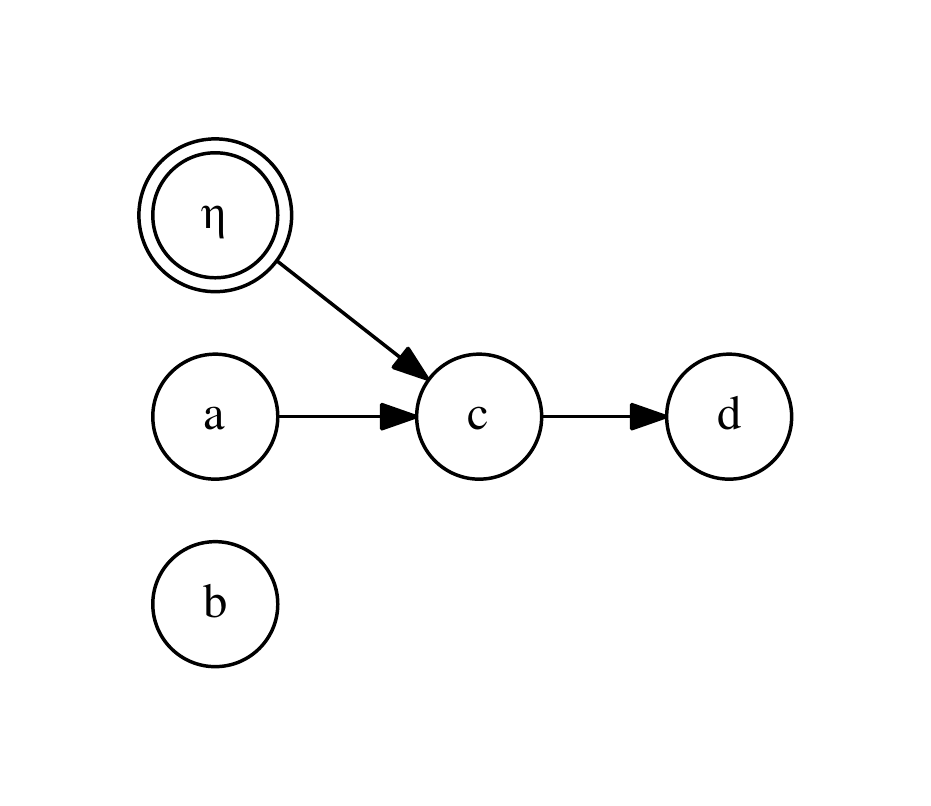}} & $(a \rightarrow c) \land \lnot (b \rightarrow c) \land (\eta \rightarrow c) \equiv (a \rightarrow c) \land \lnot (b \rightarrow c) \land \lnot c $                   & 0.054 & $\{\eta\}, \{\eta,a\}, \{\eta,b\}, \{\eta,d\}, \{\eta,a,b\},$ $\{\eta,a,d\}, \{\eta,b,d\}, \{\eta,a,b,d\}$\\
\raisebox{-0.5\totalheight}{\includegraphics[trim=35 30 35 30,clip,width=0.16\textwidth]{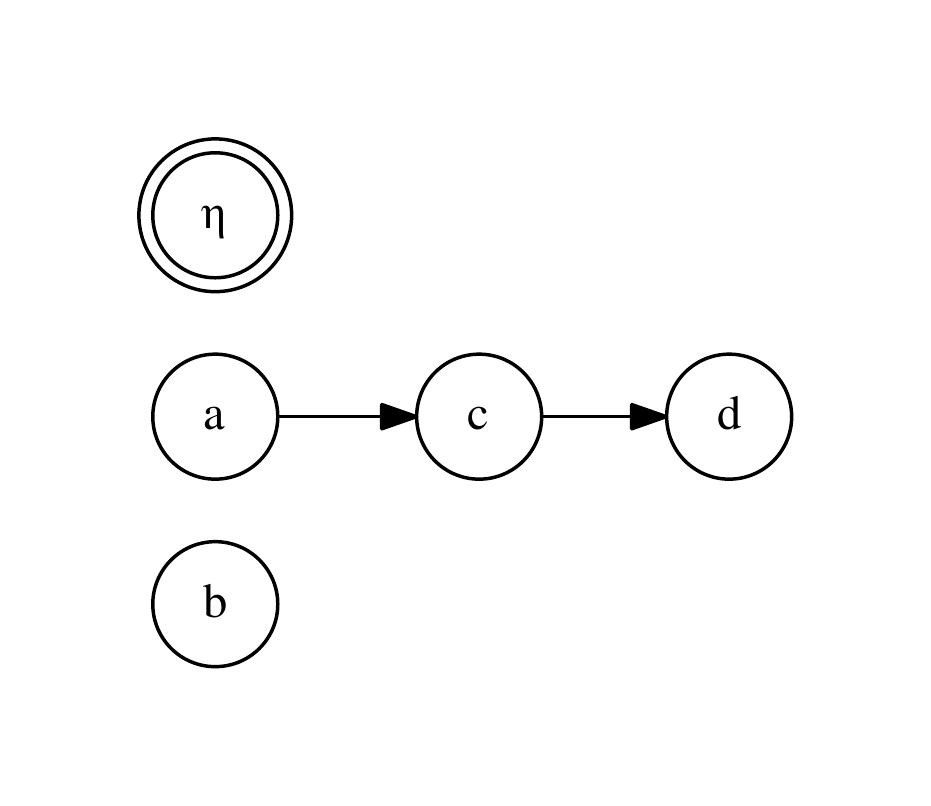}} & $(a \rightarrow c) \land \lnot (b \rightarrow c) \land \lnot (\eta \rightarrow c) \equiv (a \rightarrow c) \land \lnot (b \rightarrow c) \land c  $       & 0.036 & $\{\eta\}, \{\eta,a\}, \{\eta,b\}, \{\eta,a,b\}, \{\eta,a,d\},$ $\{\eta,a,b,d\}$\\
\raisebox{-0.5\totalheight}{\includegraphics[trim=35 30 35 30,clip,width=0.16\textwidth]{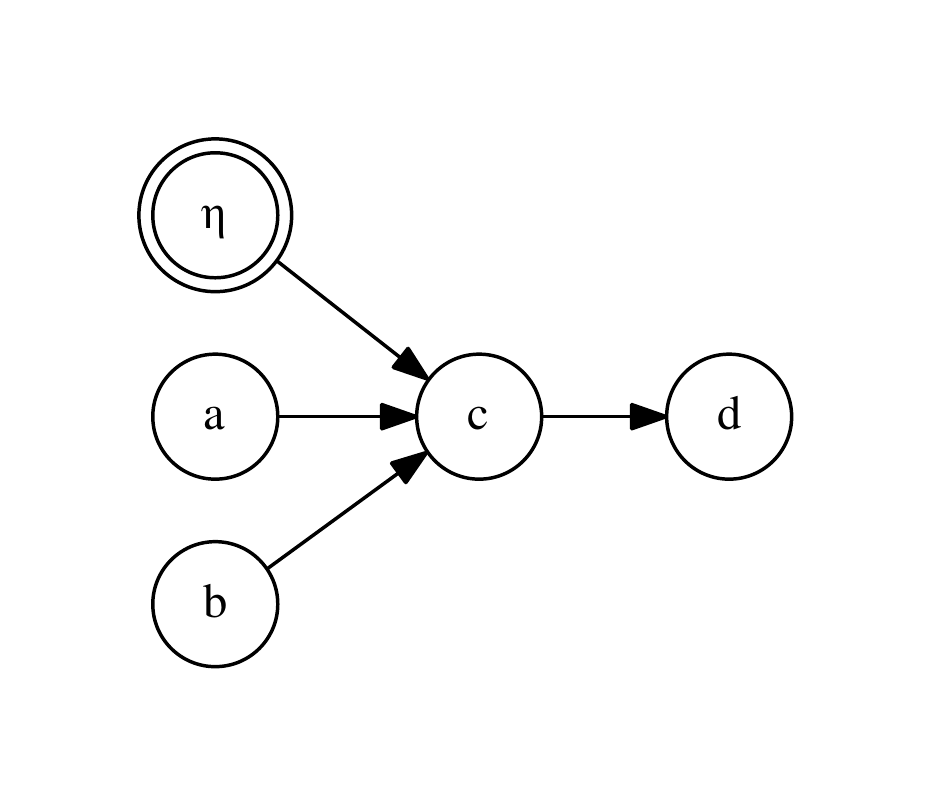}} & $(a \rightarrow c) \land (b \rightarrow c) \land (\eta \rightarrow c) \equiv (a \rightarrow c) \land (b \rightarrow c) \land \lnot c  $                   & 0.126 & $\{\eta\}, \{\eta,a\}, \{\eta,b\}, \{\eta,d\}, \{\eta,a,b\},$ $\{\eta,a,d\}, \{\eta,b,d\}, \{\eta,a,b,d\}$\\
\raisebox{-0.5\totalheight}{\includegraphics[trim=35 30 35 30,clip,width=0.16\textwidth]{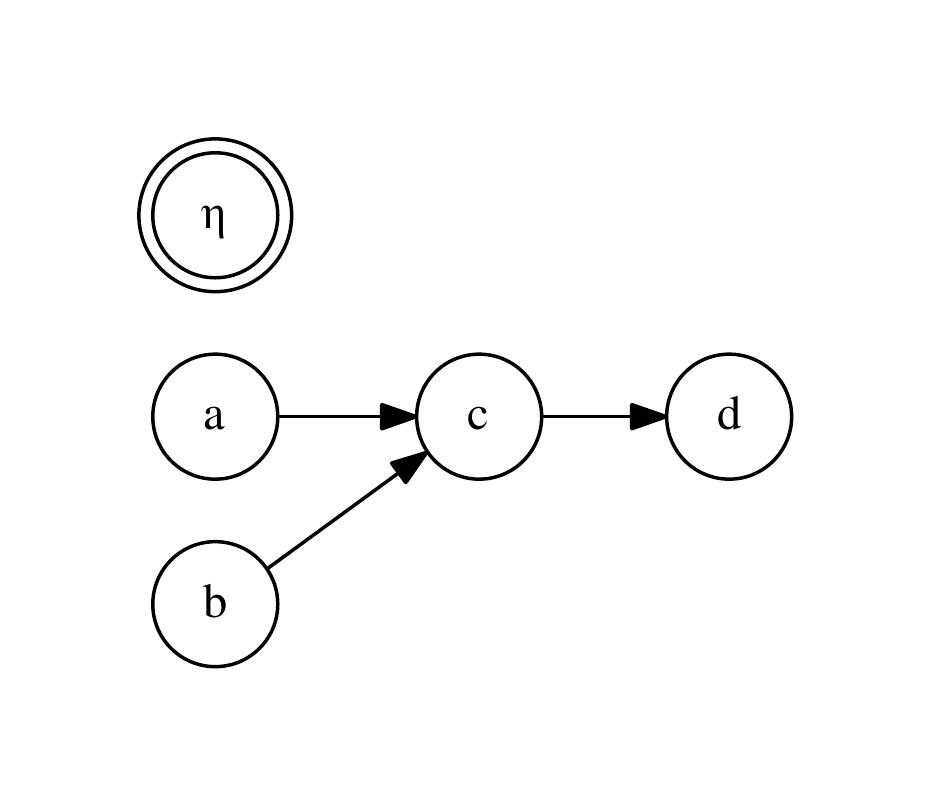}} & $(a \rightarrow c) \land (b \rightarrow c) \land \lnot (\eta \rightarrow c)  \equiv (a \rightarrow c) \land  (b \rightarrow c) \land  c $             & 0.084 & $\{\eta\}, \{\eta,a\}, \{\eta,b\}, \{\eta,a,b\}, \{\eta,a,d\},$ $\{\eta,b,d\}, \{\eta,a,b,d\}$\\\hline
\end{tabular}
\caption{Possible worlds after transforming PrAAF of Figure~\ref{fig:PrAAF}.}
\label{tbl:can_possible_worlds}
\end{table*}
\begin{proposition}[Complexity of the Transformation]
The probabilistic argument to probabilistic attack transformation has linear complexity $O(N)$ to the number of probabilistic arguments $N$ that the original PrAAF contains. It grows the size of the original PrAAF by one argument and by $N$ attacks. The transformation does not affect the worst case complexity of computing any extension or the probability that a set is any type of an extension.
\end{proposition}

\section{Conclusion and Future Work}
\label{sec:Conclusions}
In this paper we have presented a normal form for PrAAFs that is simpler but equally expressive, and a transformation that converts any general PrAAF to an equivalent PrAAF that is at a normal form. The presented normal form for PrAAFs is simpler than the general PrAAFs and allow easier modeling. With our transformation we have illustrated that the normal form can represent any general PrAAF and reproduce equal probability distributions. Our motivation with this paper is to provide a simpler but equally powerful definition for constellation PrAAFs; furthermore, we give a clear insight in the constellation semantics and its restrictions from the point of view of generating possible worlds.

For future work, we are going take advantage of the existing work from the probabilistic logic programming community to provide an efficient system that implements PrAAFs and also examine how the properties of AAF can be used to simplifying probabilistic inference in PrAAFs.


\end{document}